\documentclass[10pt,twocolumn,twoside]{IEEEtran}
\usepackage{cite}
\usepackage{amsmath,amsthm,amssymb,mathtools,bm}
\usepackage{algorithmic}
\usepackage{graphicx}
\usepackage{textcomp}
\usepackage{xcolor}

\newtheorem{assumption}{Assumption}
\newtheorem{lemma}{Lemma}
\newtheorem{theorem}{Theorem}
\newtheorem{corollary}{Corollary}
\newtheorem{remark}{Remark}

\newcommand{\R}{\mathbb R}
\newcommand{\Cspace}{\mathcal C}
\newcommand{\norm}[1]{\left\lVert#1\right\rVert}
\newcommand{\Lmat}{\mathbf L}
\newcommand{\Pmat}{\mathbf\Pi}

\def\BibTeX{{\rm B\kern-.05em{\sc i\kern-.025em b}\kern-.08em
    T\kern-.1667em\lower.7ex\hbox{E}\kern-.125emX}}
\begin{document}
\title{Directional Bounded-Confidence Cucker--Smale Flocking}
\author{Hossein B. Jond
\thanks{This work was funded by the Czech Science Foundation (GA\v{C}R) under research project no.\ 23-07517S and the European Union under the project  Robotics and Advanced Industrial Production  (reg.\ no.\ CZ.02.01.01/00/22\_008/0004590). }
\thanks{H. B. Jond is with the Department of Cybernetics, Faculty of Electrical Engineering, Czech Technical University in Prague, 121 35 Prague
2, Czech Republic (e-mail: hossein.barghi.jond@cvut.cz).}
}

\maketitle

\begin{abstract}
This paper introduces a directional bounded-confidence extension of the Cucker--Smale (CS) flocking model that decomposes pairwise relative velocities into independent radial and tangential channels within the instantaneous line-of-sight frame. Gating these components via compactly supported kernels creates a time-varying, state-dependent, matrix-weighted interaction that strictly generalizes classical scalar CS alignment. We establish an exact energy-dissipation identity with conserved mean velocity and prove exponential velocity consensus under a sliding-window persistent excitation (PE) condition formulated directly on the disagreement subspace. Furthermore, we derive a CS bounded-diameter criterion separating metric decay from aggregate directional coercivity, along with a cluster convergence guarantee. Simulations validate the exact dissipation balance and demonstrate that directional gating successfully restores global alignment in scenarios where traditional isotropic tied-confidence laws remain polarized.
\end{abstract}


\section{Introduction}
\label{sec:introduction}
The Cucker--Smale (CS) alignment model modulates influence continuously with distance to establish asymptotic velocity agreement~\cite{cucker2007,ha2008}. While various CS generalizations explore finite-speed information, delays, and topological interactions, bounded-confidence models complement these approaches by addressing information compatibility~\cite{hegselmann2002,lorenz2007}. However, standard multidimensional implementations utilize a single isotropic scalar gate on the Euclidean relative velocity. This enforces an all-or-nothing interaction that strictly couples all spatial dimensions, causing agents to abruptly ignore neighbors who share an identical heading but differ slightly in speed. 

To address this limitation and reduce undue polarization in robotic swarms, we decompose pairwise relative velocities into radial and tangential components within the instantaneous line-of-sight frame, applying independent bounded-confidence kernels to each. Radial relative motion changes range, while tangential relative motion changes bearing; because compatibility in range-rate and bearing-rate may differ, gating them independently offers a principled way to salvage partial agreement. This strictly generalizes the available edge modes to include full, radial-only, tangential-only, or rejected coupling, transforming the interaction into a positive-semidefinite matrix with state-dependent eigenvalues and eigenvectors~\cite{trinh2018}.

This physically grounded, state-dependent realization intersects with existing matrix-weighted consensus~\cite{trinh2018,miao2021,rao2025asynchronous} and bearing-based control literature~\cite{zhao2016,zhao2017}, which employ similar projector blocks (e.g., $\mathbf I-\mathbf g\mathbf g^{\mathsf T}$). Anisotropic perception has also been incorporated into CS models through vision cones and heading-dependent neighbor selection~\cite{agueh2011refined, motsch2011new, gao2026visual}. While such formulations capture directional awareness, they typically use directional perception merely to determine whether a neighbor is included, continuing to apply a scalar weight to the entire relative velocity. Our model utilizes the instantaneous line-of-sight with endogenous gating to act as an alignment filter, independently gating orthogonal directional components. This distinguishes it from both anisotropic models and bearing-based control that enforces prescribed formations. Our matrix-weighted analysis emphasizes that scalar connectivity is insufficient for systems with anisotropic sensing or trust filters: because independent gating produces potentially rank-deficient edge matrices, convergence requires the coverage of disagreement directions over time, as non-consensus directions can remain hidden within the matrix-weighted Laplacian's null space.

Consequently, our analysis builds on integral-network perspectives to overcome multidimensional communication failures. In scalar-weighted CS models, persistence-of-excitation (PE) frameworks offer a close analogue where a lower bound on time-averaged algebraic connectivity yields quantitative decay of velocity variance~\cite{bonnet2021communication,cicolani2025nonuniversal}. For time-varying matrix networks, integral-network null-space and positive spanning tree conditions provide a complementary perspective~\cite{pan2021consensus,pan2022cluster}. Compared to these geometric characterizations, our proposed PE theorem is a distinct but compatible spectral window inequality formulated directly on the disagreement subspace. While state-dependent directional gates complicate the verification of PE from local thresholds—currently requiring trajectory-level verification or conservative bounds—the resulting flocking theorems provide a structured path to certify convergence in practice. 

The primary contributions of this paper are a directional bounded-confidence CS model extension utilizing smooth compactly supported radial and tangential kernels, and the derivation of exact conservation and dissipation identities for its matrix-weighted Laplacian. We establish convergence theorems under fixed null spaces, instantaneous coercivity, and a precisely stated sliding-window excitation condition formulated on the disagreement subspace. Furthermore, we derive a CS bounded-diameter criterion that separates distance decay from aggregate directional coercivity, and we present simulations against tied-confidence, scalar anisotropy, and fixed bearing-projector baselines to underscore the necessity of state-dependent directions for achieving full consensus.

\section{Directional Cucker--Smale Model}
Consider $n$ agents in $\R^d$, each with position $\mathbf{p}_i$ and velocity $\mathbf{v}_i$. Let $\mathbf r_{ij}=\mathbf p_j-\mathbf p_i$, $d_{ij}=\norm{\mathbf r_{ij}}$, and $\mathbf w_{ij}=\mathbf v_j-\mathbf v_i$. The general alignment dynamics are governed by
\begin{equation}
 \dot{\mathbf p}_i=\mathbf v_i,\qquad
 \dot{\mathbf v}_i=\frac1n\sum_{j=1}^n
 \mathbf A_{ij}(\mathbf p,\mathbf v)\mathbf w_{ij}.
 \label{eq:model}
\end{equation}

On the collision-free set $d_{ij}>0$, we isolate the radial and tangential directions using the line-of-sight vector $\mathbf e_{ij}=\mathbf r_{ij}/d_{ij}$ and its associated orthogonal projectors:
\begin{equation}
\mathbf P_{ij}^{r}=\mathbf e_{ij}\mathbf e_{ij}^{\mathsf T},\qquad
\mathbf P_{ij}^{t}=\mathbf I-\mathbf P_{ij}^{r}.
\label{eq:projectors}
\end{equation}
The distance-based influence relies on the standard CS kernel $\phi(d)=(1+d^2)^{-\beta}$ for $\beta\ge0$.

\begin{figure*}[t]
\centering
\includegraphics[width=\textwidth]
{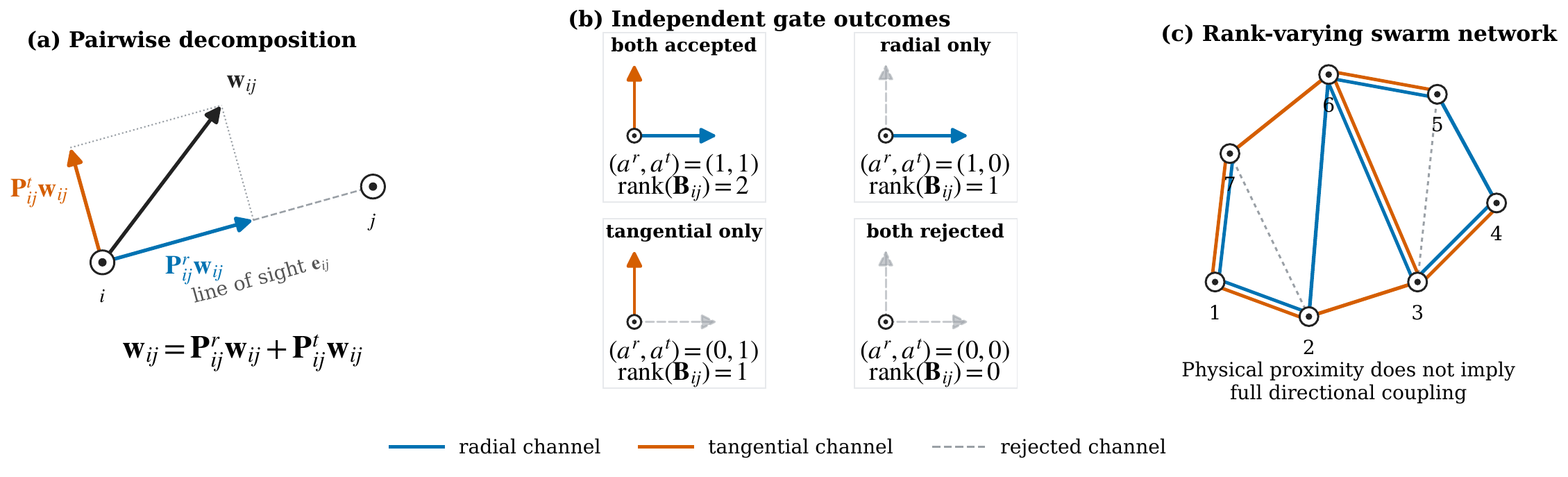}
\caption{Directional bounded-confidence mechanism. (a) Pairwise relative velocity decomposition in the instantaneous line-of-sight frame. (b) Independent gating yielding full, radial-only, tangential-only, or rejected alignment interactions. (c) The resulting state-dependent, matrix-weighted alignment network. Active radial and tangential channels are blue and orange, respectively; dashed gray denotes rejected channels.}
\label{fig:directional-confidence}
\end{figure*}

The relative velocity decomposes into radial (range-rate) and tangential (bearing-rate) components as $\mathbf{w}_{ij}=\mathbf{P}_{ij}^{r}\mathbf{w}_{ij}+\mathbf{P}_{ij}^{t}\mathbf{w}_{ij}$, as illustrated in Fig.~\ref{fig:directional-confidence}(a). We gate each component independently using a $C^1$ compactly supported confidence kernel. For $q\in\{r,t\}$ with threshold $\epsilon_q>0$, let
\begin{equation}
h_{\epsilon_q}(s)=
\begin{cases}
\left(1-s^2/\epsilon_q^2\right)^2,&0\le s<\epsilon_q,\\
0,&s\ge\epsilon_q.
\end{cases}
\label{eq:smooth-gate}
\end{equation}
The state-dependent interaction matrix $\mathbf A_{ij}$ is constructed by applying these independent gates:
\begin{align}
a_{ij}^{r}&=h_{\epsilon_r}\!\left(
\norm{\mathbf P_{ij}^{r}\mathbf w_{ij}}\right),&
a_{ij}^{t}&=h_{\epsilon_t}\!\left(
\norm{\mathbf P_{ij}^{t}\mathbf w_{ij}}\right),\label{eq:gates}\\
\mathbf B_{ij}&=a_{ij}^{r}\mathbf P_{ij}^{r} + a_{ij}^{t}\mathbf P_{ij}^{t}\succeq0,&
\mathbf A_{ij}&=\phi(d_{ij})\mathbf B_{ij}.
\label{eq:edge}
\end{align}

By construction, $\mathbf e_{ji}=-\mathbf e_{ij}$ and $\mathbf w_{ji}=-\mathbf w_{ij}$, yielding symmetric, positive-semidefinite edge weights $\mathbf A_{ij}=\mathbf A_{ji}\succeq0$. The matrix-weighted edge block $\mathbf B_{ij}$ strictly generalizes classical alignment: it permits full coupling ($a_{ij}^r>0, a_{ij}^t>0$), radial-only coupling, tangential-only coupling, or total rejection. Consequently, physical connectivity no longer guarantees alignment coupling in every disagreement direction, as $\mathbf B_{ij}$ can be rank-deficient (Fig.~\ref{fig:directional-confidence}(b)). When all channels are fully accepted ($\mathbf{P}_{ij}^{r}+\mathbf{P}_{ij}^{t}=\mathbf{I}$), the model recovers standard unfiltered CS. Conversely, setting a single tied gate $a_{ij}^r=a_{ij}^t=h_\epsilon(\norm{\mathbf w_{ij}})$ reproduces the restrictive isotropic bounded-confidence behavior, which abruptly severs multi-dimensional connections based on a single magnitude. 

Stacking the velocities as $\mathbf v=\operatorname{col}(\mathbf v_1,\ldots,\mathbf v_n)$, the alignment dynamics become $\dot{\mathbf v}=-(1/n)\Lmat(t)\mathbf v$. The state-dependent block Laplacian is defined by $[\Lmat]_{ij}=-\mathbf A_{ij}\;(i\ne j)$ and $[\Lmat]_{ii}=\sum_{j\ne i}\mathbf A_{ij}$.
Let $\otimes$ denote the Kronecker product, $\mathbf{1}_n$ the $n$-dimensional column vector of ones, and $\mathbf{I}_k$ the $k \times k$ identity matrix. We denote the global consensus subspace and its orthogonal projector as
\[
\Cspace=\{\mathbf{1}_n\otimes\mathbf{z} : \mathbf{z}\in\R^d\},\qquad
\Pmat=\left(\mathbf{I}_n-\frac1n\mathbf{1}_n\mathbf{1}_n^{\mathsf T}\right)
\otimes\mathbf{I}_d.
\]

\section{Theoretical Analysis}

\subsection{Conservation and Dissipation}

\begin{assumption}\label{ass:collision}
The initial positions are distinct, and the considered solution remains in the collision-free set. The smooth kernel \eqref{eq:smooth-gate} is used.
\end{assumption}

\begin{lemma}\label{lem:wellposed}
Under Assumption~\ref{ass:collision}, the vector field of \eqref{eq:model}--\eqref{eq:edge} is locally Lipschitz. Hence, a unique classical solution exists on every interval where the minimum pair distance remains positive.
\end{lemma}
\begin{proof}
All maps in \eqref{eq:projectors}--\eqref{eq:edge} are locally Lipschitz away from $d_{ij}=0$. Because the compact confidence kernel and $\phi$ are bounded, the standard continuation theorem applies on compact collision-free subsets.
\end{proof}

\begin{remark}
The conditional convergence analysis is not strictly limited to the specific polynomial shape of \eqref{eq:smooth-gate}. Any alternative locally Lipschitz, compactly supported profile preserves local well-posedness, pairwise symmetry, and the exact energy identities. Relaxing the kernel to be merely continuous but non-Lipschitz (e.g., $h_\epsilon(s)=\max\{0,1-s/\epsilon\}$) would break the classical well-posedness guarantee, necessitating the analysis of generalized solutions (e.g., Filippov differential inclusions). Furthermore, kernel shape affects more than numerical rate constants: different attenuation profiles alter the closed-loop trajectories, which changes subsequent gating decisions, geometry, and accumulated excitation. Thus, replacing the kernel does not automatically preserve the PE hypothesis along the resulting solution.
\end{remark}

\begin{theorem}\label{thm:energy}
Under Assumption~\ref{ass:collision}, the mean velocity $\bar{\mathbf v}=n^{-1}\sum_i\mathbf v_i$ is constant. Moreover, the velocity variance
\begin{equation}
V=\frac1{2n}\sum_{i=1}^n\norm{\mathbf v_i-\bar{\mathbf v}}^2
=\frac1{2n}\norm{\Pmat\mathbf v}^2
\label{eq:energy}
\end{equation}
satisfies the exact dissipation identity
\begin{equation}
\dot V=-D,\qquad
D=\frac1{2n^2}\sum_{i,j}
\mathbf w_{ij}^{\mathsf T}\mathbf A_{ij}\mathbf w_{ij}\ge0.
\label{eq:dissipation}
\end{equation}
\end{theorem}
\begin{proof}
Pairwise symmetry ($\mathbf A_{ij}=\mathbf A_{ji}$) implies $\sum_i\dot{\mathbf v}_i=0$. Differentiating \eqref{eq:energy}, pairing the $(i,j)$ and $(j,i)$ terms, and using $\mathbf A_{ij}\succeq0$ yields \eqref{eq:dissipation}.
\end{proof}

\begin{remark}
This exact identity applies directly to the nominal closed-loop system. Crucially, it highlights why scalar connectivity is insufficient for velocity consensus: the dissipation rate $D$ may vanish entirely for non-consensus velocities if the disagreement directions belong to $\ker\Lmat(t)$.
\end{remark}

\subsection{Fixed Matrix-Weighted Interaction}

Let $\operatorname{proj}_{\mathcal{S}}$ denote the orthogonal projection onto a subspace $\mathcal{S}$.

\begin{theorem}\label{thm:fixed}
Suppose $\Lmat(t)\equiv\Lmat_\star=\Lmat_\star^{\mathsf T}\succeq0$. Then $\mathbf v(t)\longrightarrow\operatorname{proj}_{\ker\Lmat_\star}\mathbf v(0)$. Global velocity consensus holds for every initial velocity if and only if $\ker\Lmat_\star=\Cspace$. In that case,
\[
\norm{\Pmat\mathbf v(t)}
\le e^{-\lambda_{d+1}(\Lmat_\star)t/n}
\norm{\Pmat\mathbf v(0)}.
\]
\end{theorem}
\begin{proof}
For the time-invariant system $\dot{\mathbf v}=-(1/n)\Lmat_\star\mathbf v$, diagonalizing the symmetric matrix $\Lmat_\star$ dictates that the component of $\mathbf v(0)$ in $\ker\Lmat_\star$ is strictly conserved. Meanwhile, every orthogonal eigenmode decays exponentially at the rate $\lambda_k/n$. Because the first $d$ eigenvectors span $\Cspace$, the global consensus condition follows directly. 
\end{proof}

\subsection{Instantaneous and Persistent Coercivity}

\begin{corollary}\label{cor:gap}
If $\ker\Lmat(t)=\Cspace$ and $\lambda_{d+1}(\Lmat(t))\ge\underline\lambda>0$ for almost every $t$, then $V(t)\le V(0)e^{-2\underline\lambda t/n}$.
\end{corollary}
\begin{proof}
Applying the Rayleigh quotient bound to the Laplacian quadratic form in~\eqref{eq:dissipation} yields $\dot V\le-(2\lambda_{d+1}/n)V$. Grönwall's inequality then implies the result.
\end{proof}

The instantaneous gap condition is unnecessarily restrictive for rotating or switching directional blocks. To analyze accumulated connectivity, let $\mathbf Q\in\R^{nd\times(n-1)d}$ have orthonormal columns spanning $\Cspace^\perp$, and define the symmetric, positive-semidefinite matrix function $\mathbf H: [0,\infty) \to \R^{(n-1)d \times (n-1)d}$ as $\mathbf H(t)=(1/n)\mathbf Q^{\mathsf T}\Lmat(t)\mathbf Q$. On the disagreement subspace, the alignment dynamics form a linear time-varying system.

\begin{assumption}\label{ass:pe}
The matrix-valued function $\mathbf H(t)$ is Lebesgue measurable and bounded, satisfying $0\preceq\mathbf H(t)\preceq M\mathbf I$ for some constant $M>0$ almost everywhere. There exist $T>0$ and $\alpha>0$ such that, for every $t\ge0$,
\begin{equation}
\int_t^{t+T}\mathbf H(s)\,ds\succeq\alpha\mathbf I.
\label{eq:pe}
\end{equation}
\end{assumption}

\begin{theorem}\label{thm:pe}
Under Assumption~\ref{ass:pe}, every solution of $\dot{\mathbf v}=-(1/n)\Lmat(t)\mathbf v$ satisfies
\begin{equation}
\norm{\Pmat\mathbf v(t)}
\le c e^{-\gamma t}\norm{\Pmat\mathbf v(0)},
\label{eq:pe-conclusion}
\end{equation}
where valid, albeit conservative, constants are
\begin{equation}
\eta=\frac{2\alpha}{(1+MT)^2},\quad
c=(1-\eta)^{-1/2},\quad
\gamma=-\frac{\log(1-\eta)}{2T}.
\label{eq:pe-explicit-constants}
\end{equation}
All velocities converge exponentially to the conserved mean velocity $\bar{\mathbf v}$.
\end{theorem}

\begin{proof}
Set $\mathbf z=\mathbf Q^{\mathsf T}\mathbf v$. Then $\dot{\mathbf z}=-\mathbf H(t)\mathbf z$ and $\frac{d}{dt}\norm{\mathbf z}^2=-2\mathbf z^{\mathsf T}\mathbf H\mathbf z$. For a window $[t,t+T]$, define $D_t=\int_t^{t+T}\mathbf z(s)^{\mathsf T}\mathbf H(s)\mathbf z(s)\,ds$. Since $\mathbf H^2\preceq M\mathbf H$, applying the Cauchy--Schwarz inequality yields, for every $s\in[t,t+T]$,
\[
\norm{\mathbf z(s)-\mathbf z(t)}^2
\le T\int_t^{t+T}\norm{\dot{\mathbf z}(u)}^2\,du
\le MT D_t.
\]
Since $\mathbf H(s) \preceq M\mathbf I$, this consequently bounds the weighted variation: $\int_t^{t+T}\norm{\mathbf H(s)^{1/2}(\mathbf z(t)-\mathbf z(s))}^2\,ds\le M^2T^2D_t$. Applying \eqref{eq:pe} to the fixed vector $\mathbf z(t)$ and using Minkowski's inequality in the weighted $L_2$ norm yields
\begin{align*}
\sqrt{\alpha}\,\norm{\mathbf z(t)}
&\le
\left(\int_t^{t+T}
\norm{\mathbf H(s)^{1/2}\mathbf z(t)}^2\,ds\right)^{1/2}
\\&\le(1+MT)\sqrt{D_t}.  
\end{align*}
The dissipation identity therefore gives $\norm{\mathbf z(t+T)}^2=\norm{\mathbf z(t)}^2-2D_t\le(1-\eta)\norm{\mathbf z(t)}^2$. The bounds on $\mathbf H$ imply $\alpha\le MT$, hence $0<\eta\le1/2$. Writing $\rho=\sqrt{1-\eta}$, iteration over consecutive windows and nonexpansiveness between their endpoints give
\[
\norm{\mathbf z(t)}
\le\rho^{\lfloor t/T\rfloor}\norm{\mathbf z(0)}
\le\rho^{-1}e^{(\log\rho)t/T}\norm{\mathbf z(0)}.
\]
This proves \eqref{eq:pe-conclusion} with the constants \eqref{eq:pe-explicit-constants}, since $\norm{\mathbf z}=\norm{\Pmat\mathbf v}$.
\end{proof}

\begin{remark}
The verifiable window statistic is
\begin{equation}
\lambda_{\rm PE}(t;T)=\lambda_{\min}\!\left(\int_t^{t+T}\mathbf H(s)\,ds\right).
\label{eq:lambda-pe}
\end{equation}
Theorem~\ref{thm:pe} states the rigorous implication that $\inf_{t\ge0}\lambda_{\rm PE}(t;T)>0$ guarantees exponential consensus. Furthermore, the distance kernel $\phi$ acts purely multiplicatively in $\Lmat(t)$. If inter-agent distances remain bounded over a window, $\phi(d_{ij})$ is bounded away from zero, simplifying \eqref{eq:pe} to a persistent rank condition acting solely on the evolution of the directional confidence blocks $\mathbf B_{ij}$.
\end{remark}

\subsection{Cucker--Smale Bounded-Diameter Criterion}

Let $R(t)=\norm{\Pmat\mathbf p(t)}$ and $W(t)=\norm{\Pmat\mathbf v(t)}$. By the Cauchy--Schwarz inequality, $\dot R \le W$ almost everywhere.

\begin{theorem}\label{thm:cs}
Suppose a constant $\kappa>0$ exists such that along the solution
\begin{equation}
\lambda_{d+1}(\Lmat(t))
\ge n\kappa\,\phi(2R(t)).
\label{eq:aggregate-coercivity}
\end{equation}
If
\begin{equation}
W(0)<\kappa\int_{R(0)}^\infty\phi(2s)\,ds,
\label{eq:cs-condition}
\end{equation}
then $R(t)$ remains bounded, $W(t)$ converges exponentially to zero, and all relative positions converge. For $\phi(d)$ with $\beta\le1/2$, the integral in \eqref{eq:cs-condition} diverges, making flocking unconditional whenever \eqref{eq:aggregate-coercivity} holds.
\end{theorem}

\begin{proof}
Recall the velocity variance is $V = (1/2n)W^2$. Differentiating with respect to time yields $\dot{V} = (1/n)W\dot{W}$. From the exact dissipation identity and the Rayleigh quotient bound on the consensus subspace, we established that $\dot{V} \le -\frac{2}{n}\lambda_{d+1}(\Lmat(t))V$. Substituting $V$ into this bound gives:
\begin{equation*}
\frac{1}{n}W\dot{W} \le -\frac{2}{n}\lambda_{d+1}(\Lmat(t))\left(\frac{1}{2n}W^2\right) = -\frac{1}{n^2}\lambda_{d+1}(\Lmat(t))W^2.
\end{equation*}
Applying the aggregate coercivity assumption~\eqref{eq:aggregate-coercivity}, $\lambda_{d+1}(\Lmat(t)) \ge n\kappa\,\phi(2R(t))$, we obtain:
\begin{equation*}
W\dot{W} \le -\frac{1}{n}\Big(n\kappa\,\phi(2R(t))\Big)W^2 = -\kappa\,\phi(2R(t))W^2.
\end{equation*}
For $W(t)>0$, dividing by $W$ yields the differential inequality:
\begin{equation}
\dot{W}(t) \le -\kappa\,\phi(2R(t))W(t).
\label{eq:W_dot_bound}
\end{equation}
(If $W(t)=0$, the system has reached consensus, and $\dot{W}=0$). 

We couple this with the kinematic bound $\dot{R} \le W$. Multiplying by the strictly positive term $\kappa\,\phi(2R)$ yields $\kappa\,\phi(2R)\dot{R} \le \kappa\,\phi(2R)W$. Adding this to~\eqref{eq:W_dot_bound} results in:
\begin{equation*}
\dot{W} + \kappa\,\phi(2R)\dot{R} \le 0.
\end{equation*}
Integrating this inequality over the interval $[0, t]$ and applying integration by substitution to the second term gives:
\begin{equation*}
W(t) - W(0) + \kappa\int_{R(0)}^{R(t)} \phi(2s)\,ds \le 0.
\end{equation*}
Because $W(t) = \norm{\Pmat\mathbf{v}(t)} \ge 0$, we establish an upper bound on the spatial spread:
\begin{equation*}
\kappa\int_{R(0)}^{R(t)} \phi(2s)\,ds \le W(0).
\end{equation*}
By the initial condition~\eqref{eq:cs-condition}, $W(0) < \kappa\int_{R(0)}^\infty \phi(2s)\,ds$. Since the distance kernel $\phi$ is strictly positive, the integral is a strictly increasing function of its upper limit. Thus, $R(t)$ cannot grow unbounded; there exists a finite maximum radius $R_\star \ge R(0)$ such that $R(t) \le R_\star$ for all $t \ge 0$.

Because $R(t) \le R_\star$ and $\phi$ is a nonincreasing function, the dissipation rate is uniformly bounded below by $\phi(2R(t)) \ge \phi(2R_\star) > 0$. Substituting this constant lower bound back into~\eqref{eq:W_dot_bound} yields:
\begin{equation*}
\dot{W}(t) \le -\kappa\,\phi(2R_\star)W(t).
\end{equation*}
Applying Grönwall's inequality provides the explicit exponential decay bound:
\begin{equation*}
W(t) \le W(0)e^{-\kappa\,\phi(2R_\star)t}.
\end{equation*}
Since the norm of any pairwise relative velocity satisfies $\norm{\mathbf{v}_i(t) - \mathbf{v}_j(t)} \le \sqrt{2}W(t)$, the relative velocities decay exponentially and are therefore absolutely integrable on $[0, \infty)$. Consequently, every relative position $\mathbf{p}_i(t) - \mathbf{p}_j(t) = \mathbf{p}_i(0) - \mathbf{p}_j(0) + \int_0^t (\mathbf{v}_i(s) - \mathbf{v}_j(s))ds$ converges to a finite steady-state limit.

Finally, for the standard kernel $\phi(d) = (1+d^2)^{-\beta}$, the integral $\int_{R(0)}^\infty (1+4s^2)^{-\beta}\,ds$ behaves asymptotically like $\int^\infty s^{-2\beta}\,ds$. This integral diverges to infinity if $2\beta \le 1$, meaning $\beta \le 1/2$. In this case, the right-hand side of~\eqref{eq:cs-condition} evaluates to infinity, satisfying the initial condition for any finite $W(0)$ and rendering the flocking unconditional.
\end{proof}

\begin{remark}
Condition~\eqref{eq:aggregate-coercivity} is sufficient but not necessary. Alternatively, the uniform sliding-window condition~\eqref{eq:pe} guarantees $W(t)\le c e^{-\gamma t}W(0)$, yielding $R(t)\le R(0)+cW(0)/\gamma$. Thus, uniform PE independently implies bounded relative positions and exponential velocity alignment, even if the instantaneous spectral gap vanishes.
\end{remark}

\begin{corollary}
\label{cor:design-pe}
Consider a solution satisfying the standing assumptions with the confidence kernel~\eqref{eq:smooth-gate}. Define $R_0=\norm{\Pmat\mathbf p(0)}$, $W_0=\norm{\Pmat\mathbf v(0)}$, and $\epsilon_{\min}=\min\{\epsilon_r,\epsilon_t\}$. Suppose $\sqrt{2}W_0<\epsilon_{\min}$, $m:=\left(1-2W_0^2/\epsilon_{\min}^2\right)^2$, and
\begin{equation}
W_0<m\int_{R_0}^{\infty}\phi(2s)\,ds.
\label{eq:design-flocking}
\end{equation}
Let $R_\star\ge R_0$ satisfy $m\int_{R_0}^{R_\star}\phi(2s)\,ds=W_0$. Then $R(t)\le R_\star$ and, for every $T>0$,
\begin{equation}
\lambda_{\rm PE}(t;T)\ge Tm\phi(2R_\star)>0
\qquad\text{for all }t\ge0.
\label{eq:design-pe-bound}
\end{equation}
Consequently, exponential velocity alignment and convergence of all relative positions hold. For $\phi(d)$ with $\beta\le1/2$, condition~\eqref{eq:design-flocking} is automatic.
\end{corollary}

\begin{proof}
By the exact dissipation identity (Theorem~\ref{thm:energy}), the velocity variance is nonincreasing, meaning $W(t) \le W(0) = W_0$ for all $t \ge 0$. We can bound the pairwise relative velocity using the deviation from the mean:
\begin{align*}
\norm{\mathbf{w}_{ij}(t)}^2 
&= \norm{(\mathbf{v}_j(t) - \bar{\mathbf{v}}) - (\mathbf{v}_i(t) - \bar{\mathbf{v}})}^2 \\
&\le 2\norm{\mathbf{v}_j(t) - \bar{\mathbf{v}}}^2 + 2\norm{\mathbf{v}_i(t) - \bar{\mathbf{v}}}^2 \\
&\le 2 \sum_{k=1}^n \norm{\mathbf{v}_k(t) - \bar{\mathbf{v}}}^2 = 2W(t)^2 \le 2W_0^2.
\end{align*}
Thus, $\norm{\mathbf{w}_{ij}(t)} \le \sqrt{2}W_0$ for all pairs. Because $\mathbf{P}_{ij}^r$ and $\mathbf{P}_{ij}^t$ are orthogonal projectors, they cannot increase the vector norm, ensuring $\norm{\mathbf{P}_{ij}^q \mathbf{w}_{ij}} \le \norm{\mathbf{w}_{ij}} \le \sqrt{2}W_0 < \epsilon_{\min}$ for $q \in \{r, t\}$. 

Since the confidence kernel $h_{\epsilon_q}(s)$ is strictly decreasing for $s \in [0, \epsilon_q)$, we can evaluate it at this uniform upper bound to establish a strictly positive lower bound for all state-dependent gates:
\begin{equation*}
a_{ij}^q \ge h_{\epsilon_q}(\sqrt{2}W_0) \ge \left(1 - \frac{2W_0^2}{\epsilon_{\min}^2}\right)^2 = m > 0.
\end{equation*}
This uniformly bounds the matrix-weighted edge blocks. Because $\mathbf{P}_{ij}^r + \mathbf{P}_{ij}^t = \mathbf{I}_d$, we have:
\begin{equation*}
\mathbf{B}_{ij} = a_{ij}^r\mathbf{P}_{ij}^r + a_{ij}^t\mathbf{P}_{ij}^t \succeq m(\mathbf{P}_{ij}^r + \mathbf{P}_{ij}^t) = m\mathbf{I}_d.
\end{equation*}

Similarly, spatial separation is bounded by the projected position norm: $\norm{\mathbf{p}_i(t) - \mathbf{p}_j(t)} \le \sqrt{2}R(t) \le 2R(t)$. Because the distance kernel $\phi$ is nonincreasing, $\phi(d_{ij}(t)) \ge \phi(2R(t))$. Consequently, the full interaction blocks satisfy $\mathbf{A}_{ij} \succeq m\phi(2R(t))\mathbf{I}_d$. 

The system's block Laplacian $\Lmat(t)$ is therefore bounded below by the Laplacian of a complete graph with uniform scalar edge weights $m\phi(2R(t))$. The Laplacian of such a complete graph is $m\phi(2R(t))(n\mathbf{I}_n - \mathbf{1}_n\mathbf{1}_n^{\mathsf{T}}) \otimes \mathbf{I}_d$, which equals $n m \phi(2R(t))\Pmat$. Therefore:
\begin{equation*}
\Lmat(t) \succeq nm\phi(2R(t))\Pmat.
\end{equation*}
Since $\Pmat$ is the identity operator on the disagreement subspace $\Cspace^\perp$, this implies $\lambda_{d+1}(\Lmat(t)) \ge nm\phi(2R(t))$. This satisfies the aggregate coercivity condition of Theorem~\ref{thm:cs} with $\kappa = m$. 

Applying the comparison argument from Theorem~\ref{thm:cs}, the initial condition~\eqref{eq:design-flocking} ensures that $R(t)$ is bounded by $R_\star$, where $R_\star$ solves $m\int_{R_0}^{R_\star}\phi(2s)\,ds = W_0$. Consequently, the distance kernel is uniformly bounded below by $\phi(2R_\star)$, meaning the disagreement-subspace matrix satisfies:
\begin{equation*}
\mathbf{H}(t) = \frac{1}{n}\mathbf{Q}^{\mathsf{T}}\Lmat(t)\mathbf{Q} \succeq m\phi(2R_\star)\mathbf{I}.
\end{equation*}
Integrating this constant lower bound over any window of length $T$ directly yields the persistent excitation bound~\eqref{eq:design-pe-bound}. As shown in Theorem~\ref{thm:cs}, for $\beta \le 1/2$, the integral diverges, rendering condition~\eqref{eq:design-flocking} automatically satisfied for any finite $W_0$.
\end{proof}

\begin{remark}
Corollary~\ref{cor:design-pe} provides an explicit, albeit conservative, PE lower bound linking confidence thresholds, initial spreads, and the metric kernel. Developing less restrictive conditions that exploit complementary radial and tangential activation remains an open design problem.
\end{remark}

\subsection{Cluster Convergence}

\begin{corollary}\label{cor:clusters}
Suppose the interaction graph decomposes into components $\mathcal V_1,\ldots,\mathcal V_K$ in finite time, with zero intercomponent blocks. If the restriction of \eqref{eq:pe} holds on the disagreement subspace of every component, agents in each $\mathcal V_k$ converge exponentially to that component's mean velocity. Distinct component means need not agree.
\end{corollary}

\begin{proof}
Let $T^\star < \infty$ denote the finite time after which all intercomponent interaction blocks $\mathbf{A}_{ij}(t)$ become identically zero. For all $t \ge T^\star$, the global block Laplacian $\Lmat(t)$ becomes block-diagonal (up to a permutation of agent indices), taking the form $\Lmat(t) = \operatorname{diag}(\Lmat_1(t), \ldots, \Lmat_K(t))$, where each $\Lmat_k(t)$ is the local block Laplacian for the component $\mathcal{V}_k$. 

Consequently, for $t \ge T^\star$, the global system dynamics perfectly decouple into $K$ independent subsystems:
\begin{equation*}
\dot{\mathbf{v}}_k(t) = -\frac{1}{n}\Lmat_k(t)\mathbf{v}_k(t), \quad \forall k \in \{1, \ldots, K\},
\end{equation*}
where $\mathbf{v}_k(t)$ is the stacked velocity vector of the agents in $\mathcal{V}_k$. 

Because each isolated component graph is structurally identical to the original system, each local Laplacian $\Lmat_k(t)$ retains the zero block-row sum property. Therefore, the exact dissipation identity applies locally to each component. Specifically, the local mean velocity $\bar{\mathbf{v}}_k = (1/|\mathcal{V}_k|)\sum_{i \in \mathcal{V}_k} \mathbf{v}_i(t)$ satisfies $\dot{\bar{\mathbf{v}}}_k = \mathbf{0}$ for all $t \ge T^\star$, meaning each decoupled component possesses its own strictly conserved local momentum.

By hypothesis, the persistent excitation condition~\eqref{eq:pe} holds on the local disagreement subspace of each component $\mathcal{V}_k$. Treating each decoupled component as an independent system, Theorem~\ref{thm:pe} applies directly to the local dynamics $\dot{\mathbf{v}}_k$. Thus, the local velocity variance $\norm{\mathbf{P}_k\mathbf{v}_k(t)}$ decays exponentially to zero, guaranteeing that the velocities of all agents within $\mathcal{V}_k$ converge exponentially to their conserved local mean $\bar{\mathbf{v}}_k$.

Because the subsystems are completely independent for all $t \ge T^\star$, there is no coupling force to penalize disagreement between $\bar{\mathbf{v}}_k$ and $\bar{\mathbf{v}}_l$ for $k \neq l$. Thus, global velocity consensus is not enforced, and the distinct component means need not agree.
\end{proof}

\section{Numerical Results}

To evaluate the directional bounded-confidence mechanism, we simulate ten planar agents starting from a perturbed ring with two velocity camps. The system is integrated over $T_f=30$\,s using a fourth-order Runge--Kutta method with step size $\Delta t=0.02$\,s (random seed 7). We retain the nominal dynamics~\eqref{eq:model}, the $1/n$ scaling, and the metric kernel $\phi(d)$ with $\beta=0.25$, applying identical initial conditions across five comparison laws that alter only the edge block $\mathbf B_{ij}$:
\begin{align}
\text{unfiltered:}\quad&
\mathbf B_{ij}=\mathbf I,\label{eq:unfiltered}\\
\text{tied confidence:}\quad&
\mathbf B_{ij}=h_\epsilon(\norm{\mathbf w_{ij}})\mathbf I,\label{eq:tied}\\
\text{scalar anisotropy:}\quad&
\mathbf B_{ij}=
\exp\!\left[-\frac{\norm{\mathbf P_{ij}^{r}\mathbf w_{ij}}^2}{\sigma^2}\right]
\mathbf I,\label{eq:scalar}\\
\text{fixed projector:}\quad&
\mathbf B_{ij}=\mathbf I-
\mathbf g_{ij}^{0}(\mathbf g_{ij}^{0})^{\mathsf T}.
\label{eq:fixed-projector}
\end{align}
Here, $\mathbf g_{ij}^{0}=\mathbf r_{ij}(0)/\norm{\mathbf r_{ij}(0)}$. The scalar anisotropic law~\eqref{eq:scalar} is smooth and bearing-dependent but scales the complete relative velocity uniformly. The fixed-projector law~\eqref{eq:fixed-projector} adapts the matrix block used in bearing control, albeit without a prescribed-formation objective. In contrast, only the proposed directional law~\eqref{eq:edge} independently gates the eigenvalues of the two orthogonal components.

We measure performance via the final number of connected velocity components $K_v$ (evaluated using a $0.05$\,m/s connectivity radius) and the global velocity disagreement
\[
 E_v(t)=\sqrt{\frac1n\sum_{i=1}^{n}\norm{\mathbf{v}_i(t)-\bar{\mathbf{v}}(t)}^2},
\]
defining successful alignment as $E_v(T_f)\le0.02$\,m/s. For the proposed law, we sweep the directional bounds $\epsilon_r, \epsilon_t \in \{0.30,0.55,0.80,1.10\}$\,m/s. The baseline comparisons utilize $\epsilon=0.80$\,m/s and $\sigma=0.60$\,m/s.

\begin{figure}[t]
\centering
\includegraphics[width=0.5\textwidth]{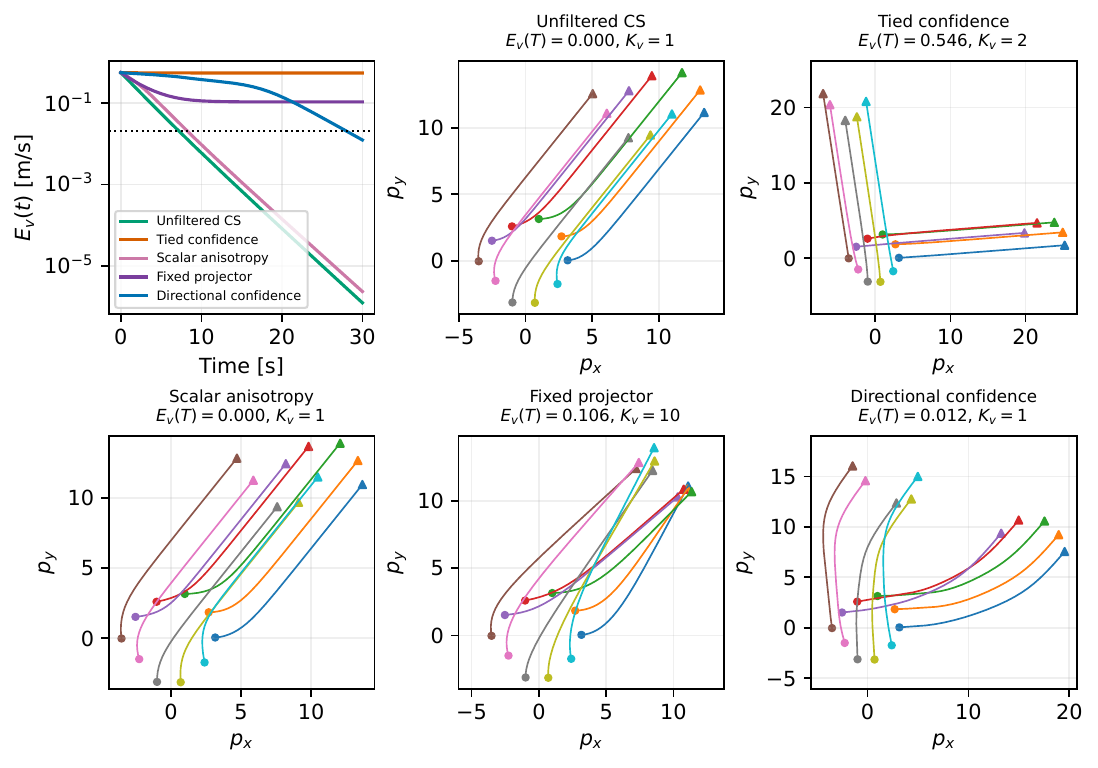}
\caption{Performance comparison of the five alignment laws under identical CS dynamics and initial conditions. Circles and triangles indicate initial and final spatial positions, respectively. The first panel tracks the global velocity disagreement $E_v(t)$, with the dotted line marking the alignment tolerance.}
\label{fig:compact}
\end{figure}

As shown in Fig.~\ref{fig:compact}, unfiltered CS and smooth scalar anisotropy align successfully, reaching $E_v(T_f)=1.24\times10^{-6}$ and $2.34\times10^{-6}$\,m/s, respectively. Tied confidence remains split ($E_v(T_f)=0.546$\,m/s, $K_v=2$) because the intercamp Euclidean velocity gap rejects the complete edge. Conversely, operating at the identical numerical bound ($\epsilon_r = \epsilon_t = 0.80$\,m/s), the proposed independent directional confidence retains compatible projected information and successfully aligns the swarm ($E_v(T_f)=0.0123$\,m/s, $K_v=1$). Finally, the fixed-projector law retains a non-consensus null direction and fails to globally align ($E_v(T_f)=0.106$\,m/s).

\begin{figure}[t]
\centering
\includegraphics[width=0.75\columnwidth]{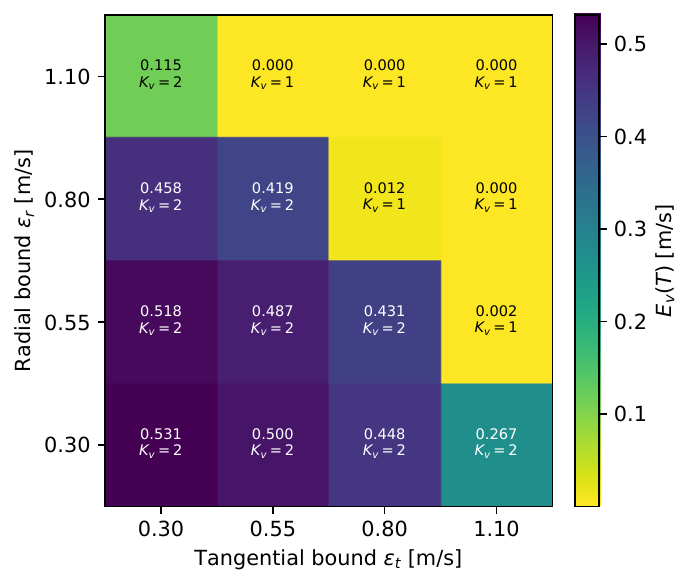}
\caption{Terminal velocity disagreement and final component count across the $4 \times 4$ grid of radial and tangential confidence bounds. The phase diagram illustrates that consensus requires sufficient complementary activation across both directional channels.}
\label{fig:phase}
\end{figure}

\begin{figure}[t]
\centering
\includegraphics[width=0.5\textwidth]{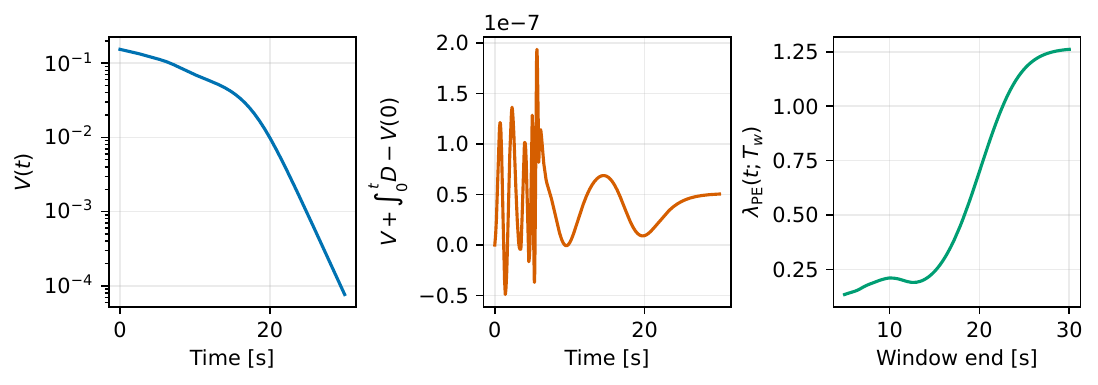}
\caption{Verification of Theorems~\ref{thm:energy} and \ref{thm:pe} for the representative directional setting. Left: Velocity variance $V(t)$. Center: Energy-balance numerical residual for $V(t)+\int_0^tD(s)ds=V(0)$. Right: Sliding-window PE statistic \eqref{eq:lambda-pe} evaluated with $T=5$\,s.}
\label{fig:diagnostics}
\end{figure}

Fig.~\ref{fig:phase} isolates the roles of the radial and tangential bounds, demonstrating that small confidence in either channel can preserve two velocity components, whereas sufficiently broad complementary channels recover matrix coercivity and consensus. Six of the sixteen tested pairs align, confirming that neither the number of scalar edges nor a single channel alone determines convergence. Fig.~\ref{fig:diagnostics} verifies the theoretical analysis: mean-velocity drift remains negligible ($<7\times10^{-16}$\,m/s), $V(t)$ is strictly monotone with an energy-balance residual of $1.94\times10^{-7}$, and the sliding-window statistic $\lambda_{\rm PE}(t;5)$ remains positive across every sampled window of the trajectory.

\subsection{Rates, Directional Modes, and Kernel Sensitivity}

\begin{figure}[t]
\centering
\includegraphics[width=0.5\textwidth]{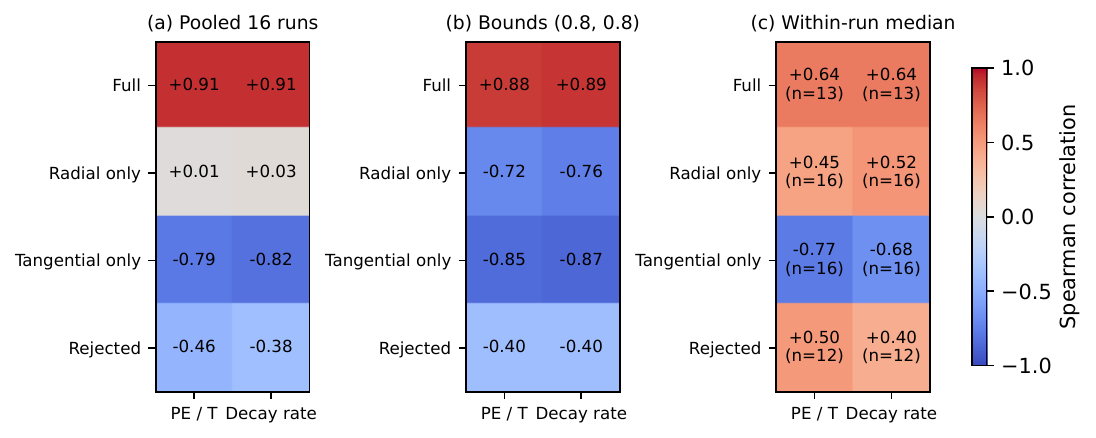}
\caption{Spearman correlations linking matching-window mode fractions to PE excitation $\lambda_{\rm PE}$ and measured RMS decay rates $\rho_E$. (a) Pooled windows across all sixteen directional confidence settings. (b) The representative setting. (c) Median within-run coefficients, where $n$ denotes the number of runs with defined correlations.}
\label{fig:mode-correlations}
\end{figure}

To relate excitation to contraction, we compare $\lambda_{\rm PE}$ with the measured root-mean-square (RMS) decay rate 
\begin{equation*}
\rho_E(t;T)=\frac1T\log\frac{E_v(t)}{E_v(t+T)},
\end{equation*}
over $5$-second windows. As shown in Fig.~\ref{fig:mode-correlations}, we compare the excitation bound $\lambda_{\rm PE}$ directly against this measured rate $\rho_E$. For the representative setting ($0.80$\,m/s), $\rho_E$ ranges from $0.0295$ to $0.252$\,s$^{-1}$, and $\lambda_{\rm PE}/T$ ranges from $0.0270$ to $0.252$\,s$^{-1}$. Averaging edge modes over all pairs and time reveals that the swarm spends $89.73\%$ in full mode, $6.07\%$ in radial-only, $4.13\%$ in tangential-only, and $0.07\%$ rejected. Across all pooled windows, full-mode occupancy exhibits strong Spearman correlations ($0.907$) with both excitation and measured rates, whereas tangential-only mode strongly anti-correlates ($-0.788$ and $-0.823$, respectively). 

\begin{figure}[t]
\centering
\includegraphics[width=0.5\textwidth]{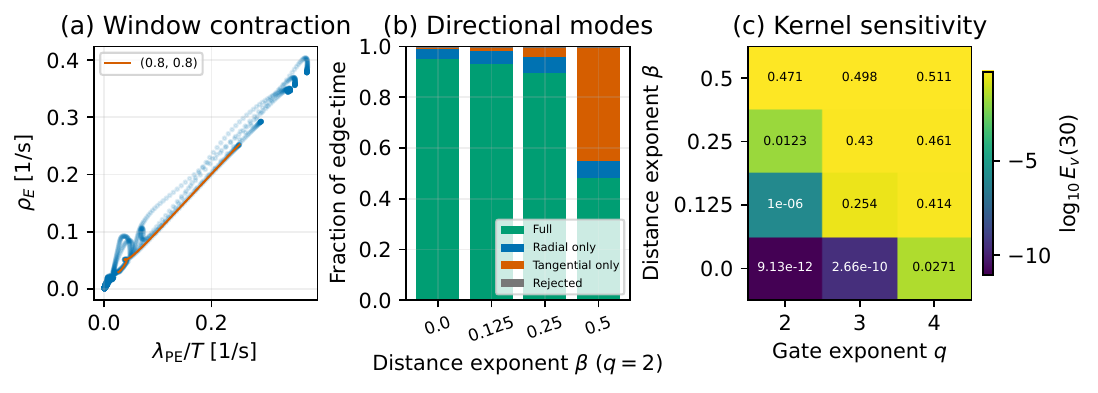}
\caption{Extended empirical diagnostics. (a) Measured RMS decay rate $\rho_E$ versus normalized PE excitation $\lambda_{\rm PE}/T$ across the confidence grid, with the representative trajectory highlighted. (b) Edge-time fractions per directional mode for $q=2$. (c) Terminal RMS disagreement across distance and confidence-kernel exponents. Panels (b) and (c) fix both confidence bounds at $0.80$\,m/s.}
\label{fig:empirical}
\end{figure}

Furthermore, the extended diagnostics in Fig.~\ref{fig:empirical} demonstrate that modifying the attenuation profile, while keeping the exact same confidence support, drastically alters the resulting trajectories, gating, and ultimate alignment. For instance, increasing the polynomial kernel exponent from $q=2$ to $q=4$ (with $\beta=0.25$) worsens terminal disagreement from $0.0123$\,m/s to $0.461$\,m/s, confirming that convergence properties are highly sensitive to the shape of the state-dependent filter.

\subsection{Explicit Cluster and Null-Space Example}

To illustrate Corollary~\ref{cor:clusters} and the matrix-weighted mechanism, we simulate a separate deterministic 1D collinear case embedded in the plane ($n=4$, $\beta=0$, $\epsilon_r=0.50$\,m/s, $\epsilon_t=0.30$\,m/s). The outer agents start at $\pm3$\,m with velocities $\pm1.1$\,m/s, while the inner agents start at $\pm2$\,m with velocities $\pm0.9$\,m/s. 

\begin{figure}[t]
\centering
\includegraphics[width=0.5\textwidth]{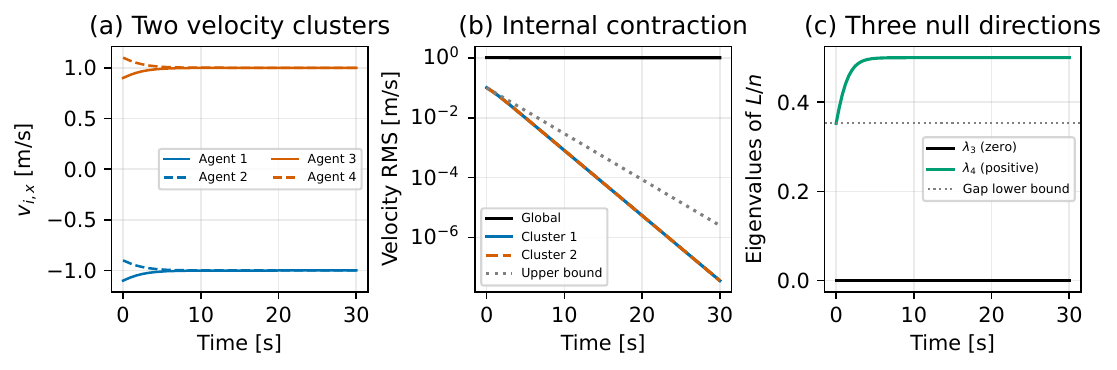}
\caption{Cluster convergence under directional gating. (a) Longitudinal velocities converging to distinct conserved pair means. (b) Within-pair RMS disagreement decaying below $0.1e^{-0.3528t}$, while global disagreement approaches $1$\,m/s. (c) Three zero eigenvalues coexisting with a uniform positive gap on the complementary subspace, illustrating how rank-one cross-pair edges preserve a non-consensus null direction.}
\label{fig:clusters}
\end{figure}

Due to the symmetry of the initial conditions, the system preserves two pairs with separating mean velocities. Every cross-pair radial relative velocity exceeds the threshold $\epsilon_r$, rendering the radial cross-pair gates identically zero; meanwhile, tangential relative velocities remain identically zero, yielding tangential gates of one. Consequently, the cross-pair interaction block is identically $\operatorname{diag}(0,1)$, while the within-pair blocks are $\operatorname{diag}(a_{ij}^r,1)$. Although the graph of nonzero matrix edges is fully connected, its block Laplacian possesses a fixed 3-dimensional kernel that preserves the longitudinal velocity difference between the clusters. As shown in Fig.~\ref{fig:clusters}, the complementary subspace contracts uniformly, driving the pairs to converge exponentially to distinct, conserved component means.

\section{Conclusion}
Integrating directional bounded confidence into the CS model establishes an endogenous alignment mechanism without relying on artificial robotic forces. By preserving pairwise symmetry, this formulation yields an exact matrix-weighted dissipation law with a conserved mean velocity, providing a foundation for comprehensive theoretical guarantees including convergence under fixed null spaces, sliding-window persistent excitation, and explicit bounded-diameter flocking criteria. 

Numerical evaluations validate these dynamics, demonstrating that independent radial and tangential gating successfully restores global alignment in scenarios where restrictive isotropic whole-velocity gating fails. Furthermore, explicit cluster simulations illustrate how rank-deficient matrix interactions can preserve non-consensus null directions even when the underlying scalar graph remains connected. 

While the PE theorem rigorously guarantees exponential consensus given a uniform window inequality, verifying this condition \textit{a priori} presents a significant challenge. Because state-dependent directional gates complicate the distance-based reductions typically used in scalar-weighted CS models, our analysis currently relies on conservative \textit{a priori} bounds based on initial spatial and velocity spreads. Furthermore, practical extensions such as collision avoidance, communication delays, and directed sensing will perturb the exact dissipation balance established here. Developing less restrictive geometric conditions that exploit complementary directional activation across varying edge blocks remains a key open problem for future research.



\section*{Acknowledgment} OpenAI's Codex (GPT-6-based) was used to assist with drafting, simulation code, language refinement, and restructuring the manuscript. The authors reviewed and revised the generated material and remain responsible for all technical content.

\bibliographystyle{IEEEtran}
\bibliography{references}

\end{document}